\documentclass[5p,times, twocolumn]{elsarticle}
\usepackage{amssymb}
\usepackage{amsmath}
\usepackage{tikz}
\usepackage{lineno}
\usepackage{bbold}
\usepackage{multirow}
\usepackage{xcolor, colortbl}
\usepackage{pifont}
\usepackage{graphicx}
\usepackage[colorlinks]{hyperref}
\def\sota{{state-of-the-art }}

\begin{document}

\begin{frontmatter}

\title{Feature Expansion and enhanced Compression for Class Incremental Learning}
\author[label1,label2]{Quentin Ferdinand \corref{cor1}}
\cortext[cor1]{corresponding author. Email: quentin.ferdinand@ensta-bretagne.org. This research was supported by Naval Group and the National Association of Research and Technology (ANRT).}
\author[label2,label4]{Benoit Clement}
\author[label3]{Panagiotis Papadakis}
\author[label1]{Quentin Oliveau}
\author[label2]{Gilles Le Chenadec}

\affiliation[label1]{organization={Naval Group},
            country={France}}
\affiliation[label2]{organization={ENSTA Bretagne},
            addressline={Lab-STICC UMR 6285}, 
            city={Brest},
            country={France}}
\affiliation[label3]{organization={IMT Atlantique},
            addressline={Lab-STICC UMR 6285, team RAMBO}, 
            city={Brest},
            country={France}}
\affiliation[label4]{organization={CROSSING IRL 2010},
            city={Adelaide},
            country={Australia}}
\begin{abstract}
Class incremental learning consists in training discriminative models to classify an increasing number of classes over time. However, doing so using only the newly added class data leads to the known problem of catastrophic forgetting of the previous classes. Recently, dynamic deep learning architectures have been shown to exhibit a better stability-plasticity trade-off by dynamically adding new feature extractors to the model in order to learn new classes followed by a compression step to scale the model back to its original size, thus avoiding a growing number of parameters. In this context, we propose a new algorithm that enhances the compression of previous class knowledge by cutting and mixing patches of previous class samples with the new images during compression using our Rehearsal-CutMix method. We show that this new data augmentation reduces catastrophic forgetting by specifically targeting past class information and improving its compression. Extensive experiments performed on the CIFAR and ImageNet datasets under diverse incremental learning evaluation protocols demonstrate that our approach consistently outperforms the \sota. The code will be made available upon publication of our work\footnote[1]{https://github.com/QFerdi/FECIL}.
\end{abstract}


\end{frontmatter}
\footnotetext[1]{https://github.com/QFerdi/FECIL}

\section{Introduction}
\label{sec:introduction}
In recent years, deep learning has undergone a remarkable evolution, demonstrating impressive achievements in various domains \cite{lecun_deep_2015, goodfellow_deep_2016, radford_language_nodate}. In particular, in the field of visual classification, convolutional neural networks have been shown to attain and even exceed human performance \cite{russakovsky_imagenet_2015, he_deep_2015, szegedy_inception-v4_2017, tan_efficientnet_2019}. Despite reaching human-like performance on specific vision tasks, however, these models encounter limitations in their ability to continually learn and adapt to novel concepts \cite{scheirer_toward_2013}, a capability inherent in human cognition. In fact, this inability to adapt incrementally poses significant challenges in real-world applications like face recognition\cite{ozawa_incremental_2005, madhavan_incremental_2021}, robotics\cite{thrun_lifelong_1995, lesort_continual_2020} up to autonomous driving \cite{pierre_incremental_2018}. The field of class incremental learning specifically studies the process of incrementally training a classification model on newly acquired data to accommodate new classes or concepts over time. The main challenge in class incremental learning is called catastrophic forgetting and refers to the tendency of models trained incrementally to forget previously learned classes when learning new ones, leading to a degradation in performance on earlier tasks or classes.
Constituting an open research problem, this has led to a plurality of approaches that seek to alleviate its effects. Among them, memory rehearsal~\cite{rebuffi_icarl_2017, wu_large_2019, dhar_learning_2019,liu_mnemonics_2020, zhao_maintaining_2020, douillard_small-task_2020, yan__2021, wang_foster_2022} stands out as one of the most widely adopted methods. This technique involves maintaining a fixed-size memory containing a few exemplars from previously learned classes. During model fine-tuning, this memory is incorporated alongside the new class data to retain a subset of samples from prior classes within the dataset, thus preventing complete forgetting. This rehearsal strategy leads to an initial performance gain that many approaches~\cite{castro_end--end_2018, wu_large_2019, douillard_small-task_2020, hou_learning_2019, zhao_maintaining_2020, ferdinand_attenuating_2022} seek to extend by combining it with knowledge distillation so as to further reduce forgetting. Knowledge distillation consists in transferring "knowledge" from one model to another, which in incremental learning translates to transferring knowledge from the previous model to the one being finetuned on new data. This enables the training model to preserve knowledge of previous classes while adapting to new ones, thereby mitigating forgetting.

While effective at reducing forgetting these methods also reduce the adaptation of the model and therefore induce a stability/plasticity trade-off \cite{grossberg_adaptive_2013, yan__2021} dilemma between retaining past classes and adapting to new classes. Recently, a new paradigm \cite{yan__2021, wang_foster_2022, li_preserving_2021} has emerged that proposes to freeze previous feature extractors and dynamically expand the feature space of the model by training new extractors during each incremental step. This novel technique has demonstrated a superior trade-off between stability and plasticity compared to conventional incremental methods~\cite{yan__2021}. This performance gain, however, comes at the cost of an increasing number of network parameters over time.

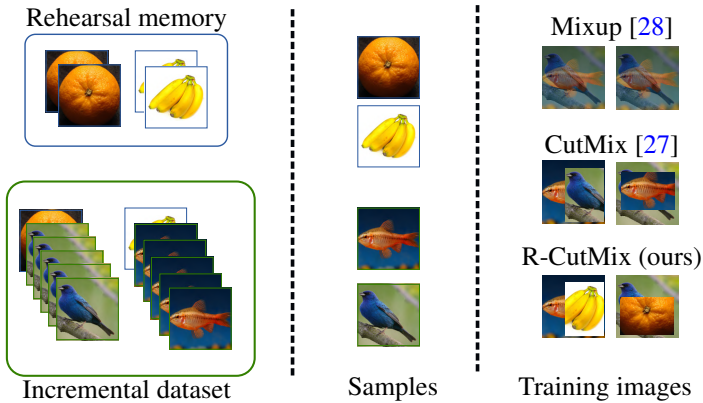
\begin{figure}[h]
    \centering
    \begin{tikzpicture}
        \node[anchor=south west,inner sep=0] (image) at (0,0) {\includegraphics[width=\linewidth]{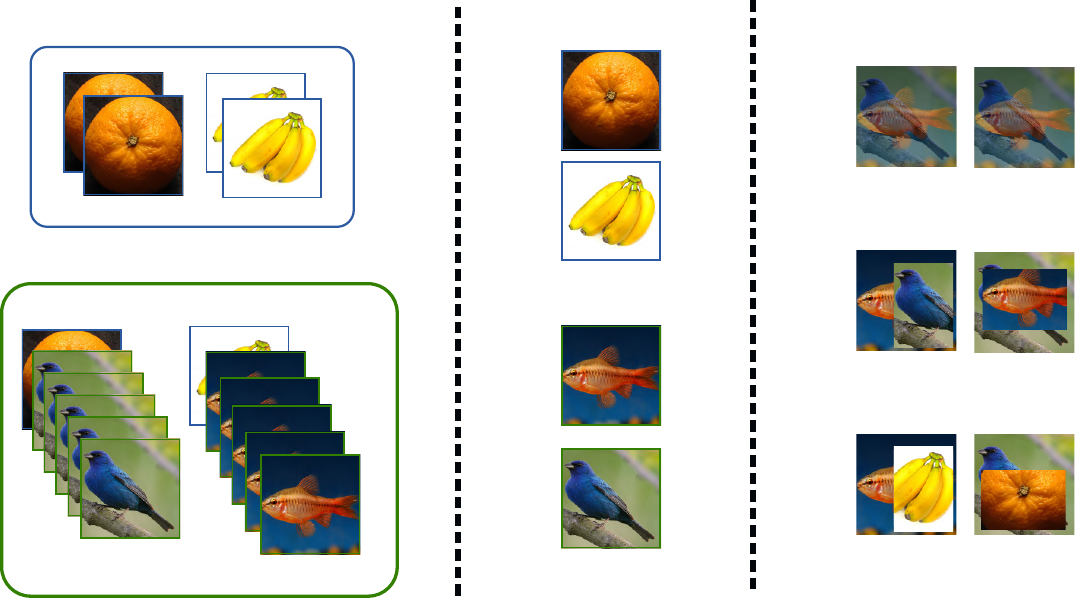}};
        \begin{scope}[x={(image.south east)},y={(image.north west)}]
            \draw (.18, .96) node {Rehearsal memory};
            \draw (.18, -0.03) node {Incremental dataset};
            \draw (.575,-0.03) node {Samples};
            \draw (.89,-0.03) node {Training images};
            \draw (.9, .935) node {Mixup \cite{zhang_mixup_2017}};
            \draw (.9, .62) node {CutMix \cite{yun_cutmix_2019}};
            \draw (.9, .32) node {R-CutMix (ours)};
        \end{scope}
    \end{tikzpicture}
    \caption{Overview of the differences between Mixup~\cite{zhang_mixup_2017}, CutMix~\cite{yun_cutmix_2019}, and our Rehearsal-CutMix procedure. Our method specifically samples one image from the incremental dataset containing mostly new classes and one image from the rehearsal memory containing only previous classes before mixing them together for training.}
    \label{fig:shema_Rcutmix}
\end{figure}

To address this issue, a plausible solution involves compressing the model back to its initial size upon the completion of each incremental step~\cite{wang_foster_2022}. This compression training step is done on the incremental dataset that is biased towards new classes and therefore impedes the compression of previous class knowledge. To address this limitation, we propose a novel method termed as \textbf{Rehearsal-CutMix} that is shown to enhance the compression step. Specifically, the key contributions of this work can be summarized as follows:
\begin{itemize}
    \item We present a novel incremental algorithm based on the expansion/compression paradigm for incremental learning.
    \item We introduce a new hybrid data augmentation technique that combines the widely adopted CutMix augmentation~\cite{yun_cutmix_2019} with the rehearsal memory employed in incremental learning as illustrated in \ref{fig:shema_Rcutmix}.
    \item Through comprehensive experimentation on multiple datasets and popular incremental evaluation protocols we demonstrate the effectiveness of this new rehearsal-CutMix augmentation when used during the compression step of our algorithm. Furthermore, our method attains the best performance against the \sota on all evaluation datasets.
\end{itemize}

This work follows up on our previous study \cite{ferdinand_attenuating_2022} that leveraged contrastive learning methods to improve the quality of the features learned incrementally. Going beyond that earlier work, we take advantage of the improved stability/plasticity trade-off of dynamic models and enhance the necessary compression step with a new data augmentation based on the popular CutMix \cite{yun_cutmix_2019} technique, recognized for its proven efficacy in enhancing model generalization. 

The remainder of the article is organized as follows. In section \ref{sec:related_works} we conduct a comprehensive review of related works in the class incremental field, providing some context to the different components of our method. In section \ref{sec:method} we then describe in detail our methodology and the integration of the Rehearsal-CutMix data augmentation technique to the compression step. The performance evaluation of our approach is unfolded in section \ref{sec:experiments}, where we present experimental results and comparative analyses across multiple datasets and methods, demonstrating the superiority of our framework. Finally, section \ref{sec:conclusion} concludes our work and identifies directions for future investigations.

\section{Related works}
\label{sec:related_works}
In this section, we first review the main approaches used to alleviate catastrophic forgetting during incremental training. Subsequently, we describe the emerging paradigm that dynamically expands the trained network and has been shown to lead to better plasticity/stability trade-offs during incremental steps. Finally, a description of the mixup data-augmentation strategy and its importance in the context of incremental learning is explored.

\subsection{Conventional methods}
Class incremental learning methods aim to train models able to learn new classes incrementally in a way that controls the forgetting of previous ones. We refer the interested reader to the following surveys~\cite{masana_class-incremental_2020, zhou_deep_2023} for a detailed presentation of the \sota, allowing us to focus on the main components in the following paragraphs.

In the past years, numerous methods have been proposed to alleviate catastrophic forgetting during incremental learning. Notably, knowledge distillation, initially introduced as a means to transfer knowledge from a larger teacher model to a smaller student model~\cite{hinton_distilling_2015}, found its early application in incremental learning through the method \textit{Learning without Forgetting}(LwF)~\cite{li_learning_2018}. By using the probabilistic model output of the previous incremental step as a soft target for the model being trained during an incremental step, Li et al. were able to transfer knowledge from previous classes into the model learning new classes, thus mitigating forgetting. 

Subsequent developments instigated by the \textit{iCarL} method~\cite{rebuffi_icarl_2017}, expanded on this basis by combining it with a rehearsal memory mechanism to further reduce forgetting~\cite{castro_end--end_2018, wu_large_2019, zhao_maintaining_2020}. This procedure consists in storing the most representative exemplars of each seen class within a fixed-size memory in order to be able to jointly replay them with new data during incremental steps. 

This approach posed a novel challenge by inducing an imbalance in the incremental dataset towards new classes that is known to be particularly problematic for the training of deep neural networks~\cite{buda_systematic_2018} and was shown to bias classification towards the most represented classes of the dataset~\cite{masana_class-incremental_2020}. In \textit{EtEIL}~\cite{castro_end--end_2018}, the authors proposed fine-tuning the classification layer on a balanced subset of the training data while \textit{DER}~\cite{yan__2021} trains a new classification layer from scratch on this subset. The alternative method of rescaling weights of new and past classes within the biased classification layer has also been explored. In \textit{Bic}~\cite{wu_large_2019}, authors apply an affine rescaling of the weights and learn its parameters on a balanced validation set. Finally, attaining comparable performance, the method \textit{WA}~\cite{zhao_maintaining_2020} greatly simplified this rebalancing step by simply rescaling the classification layer based on the norm ratio of past and new class weights.

\subsection{Dynamic models}
\label{sec:Dyna_models}

An emerging concept in incremental learning advocates freezing the learned weights after each incremental step and adding new ones for adaptation to new classes. Early approaches considered using completely different sets of neurons~\cite{fernando_pathnet_2017, hung_compacting_2019} for each incremental step but were inherently bounded by the network capacity, depleted only after training on a few incremental tasks. 
    
Towards overcoming this limitation, recent works explored the addition of new neurons~\cite{hung_compacting_2019} or even new complete feature extractors incrementally to accommodate new classes~\cite{yan__2021, li_preserving_2021, wang_foster_2022}. These approaches typically induce a considerable increase in network parameters with each incremental step that some works try to mitigate with complex pruning-based approaches~\cite{hung_compacting_2019, yan__2021, li_preserving_2021}. These approaches typically involve training models with sparsity losses to then be able to prune many useless neurons from the model in order to reduce significantly the number of parameters with a minimal loss of performance. This type of approach mitigates but does not solve completely the issue of the number of required parameters growing with the incremental steps and generally requires specific hyper-parameters depending on the dataset and application to find the right balance between performance and parameter growth.

The recent method \textit{FOSTER}~\cite{wang_foster_2022}, however, considered a concurrent approach making use of the knowledge distillation loss to compress the dynamic network with added neurons back to its original size after each incremental step. This compression necessitates a dedicated training step on the incremental dataset and therefore faces the typical challenges of incremental learning such as forgetting of past class information and training on an imbalanced dataset. 

To alleviate the previous shortcoming, in this work we introduce a novel data augmentation based on the popular \textit{CutMix} augmentation~\cite{yun_cutmix_2019} and designed specifically to improve the distillation of previous classes during this compression step.

\subsection{Mixup-based data augmentation}

\textit{Mixup}~\cite{zhang_mixup_2017} is a data augmentation technique that consists in generating and training on random interpolations between samples. Despite its simplicity this augmentation has been shown to improve significantly the generalization of \sota neural network architectures. Specifically, training classification models on interpolations between samples leads to smoother decision boundaries between classes therefore improving the generalization of the model.

Recently, many variants of this mixup procedure have been proposed \cite{yun_cutmix_2019, verma_manifold_2019, chou_remix_2020} by changing the method used to mix different samples and labels. \textit{CutMix} \cite{yun_cutmix_2019}, the main method considered in this paper, combines \textit{Mixup} with the \textit{Cutout} data augmentation \cite{devries_improved_2017} by cropping and mixing patches of different images instead of interpolating between them. This method of mixing images together was shown to retain the benefits of both \textit{Cutout} and \textit{Mixup} and further reduce the over-confidence issue that can arise when training deep neural networks.

Manifold \textit{Mixup} \cite{verma_manifold_2019} on the other hand considers interpolations between hidden representations of the samples to learn class manifolds with less variance to further improve the robustness of models trained in this way. While superior to the base \textit{Mixup} augmentation, Manifold \textit{Mixup} requires two images to go through the feature extractor part of the network before mixing them to obtain the final training sample and therefore induces an increased training time and gpu memory cost when training the model.

\section{Proposed Method}
\label{sec:method}
\begin{figure*}[!t]
	\centering
	\includegraphics[width=\textwidth]{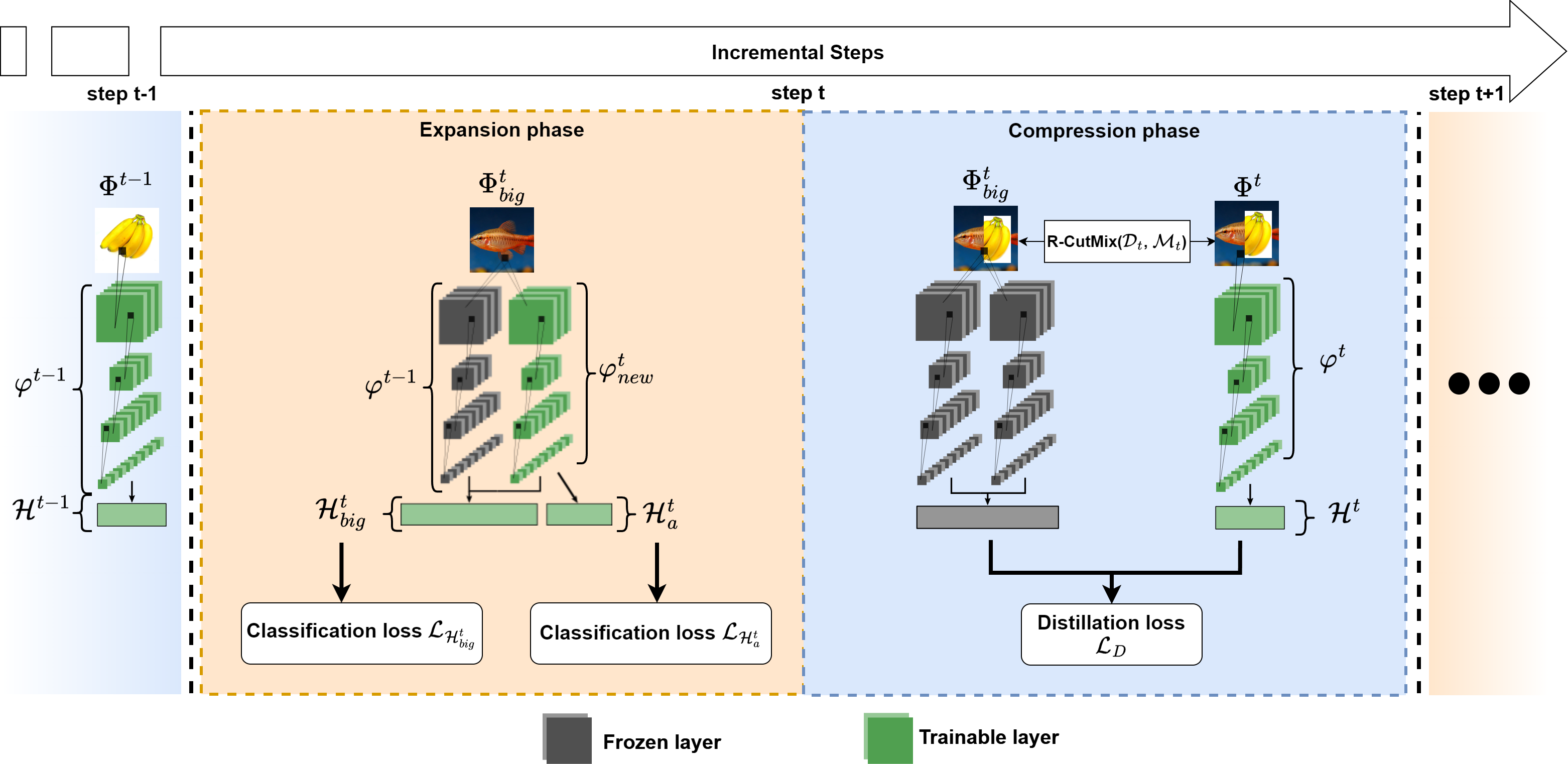}
	\caption{Pipeline of the proposed approach. Each incremental step consists of two training phases, first the expansion phase where we dynamically expand $\Phi^{t-1} $ to learn new classes, and second, the phase where we compress the expanded model $\Phi^{t}_{big}$ back to its original size with minimal performance drop using our Rehearsal-CutMix distillation mechanism.}
	\label{fig:shema_fecil}
\end{figure*}

In this section, we describe in detail our method called Feature Expansion and enhanced Compression for Incremental Learning (FECIL). A schema representing the overall pipeline is shown in Figure \ref{fig:shema_fecil} while the details of the expansion and compression stages are explained in sections \ref{subsec:expansion} and \ref{subsec:compression} respectively.

\subsection{Method overview and problem setting}

In the class incremental learning problem that we consider, a model is trained sequentially for $T$ classification tasks, wherein each task incorporates a set of $C_{new}$ new classes that the model is required to classify. For each classification task $t$, a new dataset $\mathcal{D}_{new} = \{\mathcal{X}, \mathcal{Y}\}$ is considered, where $\mathcal{X}$ and $\mathcal{Y}$ represent a set of labeled samples (images in our case) belonging to $C_{new}$ classes that were previously unknown. Our method makes use of the rehearsal strategy introduced in \cite{rebuffi_icarl_2017} that stores the most representative samples of each class using Herding sampling \cite{welling_herding_2009}. These representative samples are stored in a small fixed-size memory $\mathcal{M}_t$ for future incremental training steps which leads to an augmented training dataset $\mathcal{D}_t = \{D_{new} \bigcup \mathcal{M}_t\}$ containing a part of all $C_t$ classes including those of the previous classes.

Each incremental step is split into two training phases. The first one is the expansion phase where we create and train an expanded network $\Phi^t_{big}$ by adding a new feature extractor $\varphi^t_{new}$ and weights for the new classes in the classification layer $\mathcal{H}^t$ before updating the models' parameters on the new data task. By freezing the previous feature extractor and training a new one, it is ensured that no forgetting of previous features happens in this expansion phase, however the training dataset is still imbalanced towards new classes, therefore the prominent weight alignment (WA) technique \cite{zhao_maintaining_2020} is used after this training step to remove the bias from the classification layer.

The resulting dynamic network contains two feature extractors and one classification layer allowing to classify both new and past classes. In order to prevent the number of parameters from growing after each incremental step, this network is then compressed into a more compact network $\Phi^t$ that is then saved for the next incremental step. Knowledge distillation is a technique that has witnessed particular success in non-incremental scenarios for this kind of network compression problems \cite{hinton_distilling_2015, tung_similarity-preserving_2019, tian_contrastive_2020, gou_knowledge_2020, beyer_knowledge_2022, xu_knowledge_2020}. In the context of incremental learning, however, it has been shown by the FOSTER method \cite{wang_foster_2022} to struggle with previous classes compression due to the imbalance of the incremental dataset. We therefore introduce a new data-augmentation scheme based on the CutMix method that we call Rehearsal-CutMix, specifically designed to improve the distillation of past classes knowledge in this compression step. Finally, similarly to our expansion step, any bias resulting from the dataset imbalance is removed from the classification layer with the WA technique after this training step.

Upon compression of $\Phi^t_{big}$ into $\Phi^t$, $\Phi^t_{big}$ is then discarded and only $\Phi^t$ is retained and used in the next incremental step, which ensures the model size does not grow with the incremental steps.
	
\subsection{Feature expansion}
\label{subsec:expansion}

This phase bears similarities with the one introduced in DER \cite{yan__2021}. At incremental task $t$, like DER, we create the dynamic model $\Phi^t_{big}$ that expands the previous feature space with a new feature extractor $\varphi^t_{new}$ to accommodate new classes. Unlike DER, however, our dynamic model does not require all previous extractors but only two, $\varphi^t_{new}$ and $\varphi^{t-1}$ thanks to the compression step we will detail in section \ref{subsec:compression}. Both the previous and new feature extractors are then fed into a new classifier $\mathcal{H}_{big}^t$ with $C_t$ (the total number of old and new categories) outputs. 
 
More specifically, given an input image $x$ from the incremental dataset $\mathcal{D}_t$, the feature vector $\phi$ of the model becomes the concatenation of both feature extractor outputs: 

\begin{equation}
    \phi = \{\varphi^{t-1}(x), \varphi^t_{new}(x)\}
\end{equation}

In order to mitigate the forgetting of past classes, the previous feature extractor $\varphi^{t-1}$ as well as the statistics of its batch normalization layers \cite{ioffe_batch_2015} are frozen for this expansion training step. On the other hand, in order to encourage the model to adapt the extracted features to new classes, the weights of $\varphi^t_{new}$ are initialized with those of $\varphi^{t-1}$ but optimized along with the classification layer on the incremental dataset $\mathcal{D}_t$.

The feature vector $\phi$ is then fed into the dense classification layer $\mathcal{H}_{big}^t$ and softmax is applied to the output logits of this classifier to make the prediction for each class:

$$ p_{\mathcal{H}_{big}^t}(y|x) = \text{Softmax}(\mathcal{H}_{big}^t(x))$$

Finally, the weights of the new classifier $\mathcal{H}_{big}^t$ corresponding to the old classes are initiated with those of $\mathcal{H}_{t-1}$ to retain old knowledge while the newly added weights are randomly initialized.

\paragraph{Feature adaptation to new classes} During the expansion step, the model $\Phi^t_{big}$ is trained on the incremental dataset $\mathcal{D}_t$ to classify correctly both new and old classes by minimizing a cross entropy loss function :

\begin{equation}
    \mathcal{L}_{\mathcal{H}^t_{big}} = -\frac{1}{B} \sum_{i=1}^{B} \log (p_{\mathcal{H}_{big}^t}(y = y_i|x_i))
\end{equation}
where $B$ is the size of the batch of images sampled from $\mathcal{D}_t$, $x_i$ is one image of the batch and $y_i$ is its label.

However, training solely with this loss function would lead the new feature extractor $\varphi^t_{new}$ to learn features that discriminate between past classes. These features are not only redundant with those of the feature extractor $\varphi^{t-1}$ kept frozen but used by the model, but also tend to over-fit on the memory data, which negatively impacts performance as demonstrated in \cite{yan__2021}.

For this reason we enforce the model to learn discriminating features only for new classes by employing the auxiliary classifier $\mathcal{H}^t_a$ introduced in DER. The features from $\varphi^t_{new}$ are fed into $\mathcal{H}^t_a$ made with $C_{new}+1$ outputs in order to classify all new classes and treat all past classes as one category. Initialized randomly, this classifier is then trained with a cross-entropy loss $\mathcal{L}_{\mathcal{H}^t_a}$ :

\begin{equation}
    \mathcal{L}_{\mathcal{H}^t_{a}} = -\frac{1}{B} \sum_{i=1}^{B} \log (p_{\mathcal{H}_{a}^t}(y = \tilde{y}_i|x_i))
\end{equation}
where $\tilde{y}_i$ is the modified one-hot target vector of size $C_{new}+1$. This classifier's only purpose is to moderate the training of $\varphi^t_{new}$ and is therefore discarded at the end of the training step.

Our total loss for the expansion training phase therefore becomes the following linear combination of the previously explained losses:

$$ \mathcal{L}_{exp} = \mathcal{L}_{\mathcal{H}^t_{big}} + \mathcal{L}_{\mathcal{H}^t_{a}} $$

\subsection{Feature compression}
\label{subsec:compression}

Upon completion of the feature expansion phase, our model $\Phi^t_{big}$ composed of two feature extractors obtains excellent performances, as will be demonstrated in section \ref{sec:experiments}. This performance gain comes at the cost of an increased number of parameters that we solve by compressing $\Phi^t_{big}$ back to its original size while controlling the loss of information.

Specifically, a model $\Phi^t$ composed of only one feature extractor $\varphi^t$ and a classifier $\mathcal{H}^t$ is initialized with $\Phi^{t-1}$ and trained on $\mathcal{D}_t$ with the standard knowledge distillation loss using $\Phi^t_{big}$ probability distributions as a soft targets:

\begin{equation}
    \begin{split}
    q^{\mathcal{H}^t}_c(x) &= \frac{e^{o_c(x)/\tau}}{\sum_{i=1}^{C^t} e^{o_i(x)/\tau}}\\
    \mathcal{L}_{D}(x) &= \sum_{c=1}^{C_{t}} -q^{\mathcal{H}^t_{big}}_c(x)\ \log (q^{\mathcal{H}^t}_c(x))  
    \end{split}
\end{equation}
with $q^{\mathcal{H}^t}_c(x)$ the softened softmax probability obtained from output node $o_c$ of the model, $\tau$ a temperature hyper-parameter, and $q^{\mathcal{H}^t_{big}}_c(x)$ the equivalent softened softmax probability but obtained from the outputs of the big model $\Phi^t_{big}$.
	
Due to the imbalance of $\mathcal{D}_t$, however, such a compression scheme performs better in new than past classes \cite{wang_foster_2022}, leading to a compressed model with poor performances on previous classes. We therefore devise a new method called Rehearsal-CutMix to improve the distillation of past classes.

\paragraph{Rehearsal-CutMix}

CutMix \cite{yun_cutmix_2019} is a data augmentation strategy building upon the famous Mixup augmentation \cite{zhang_mixup_2017} by mixing patches of different samples to generate synthetic training samples. This augmentation has been shown to enhance the model generalization and robustness and alleviate its overconfidence when making predictions.

Specifically, considering an image  $x \in \mathbb{R}^{W \times H \times C}$ with $W$ representing the width, $H$ the height, and $C$ the number of channels ($3$ for RGB images); and its label $y$ from the sampled mini-batch in onehot format, CutMix generates a training sample ($\Tilde{x}$,$\Tilde{y}$) by cropping and replacing patches of one image $x_i$ with patches of another image $x_j$ :

\begin{equation}
  \label{eq:mixup_cutmix}
  \begin{split}
    \Tilde{x} &= \text{M} \odot x_i + (\mathbb{1}-\text{M}) \odot x_j \\
    \Tilde{y} &= \lambda y_i + (1-\lambda) y_j \\
  \end{split}
\end{equation}
where $\text{M} \in \{0,1\}^{W \times H \times C}$ represents a binary mask, $\mathbb{1}$ a similar mask but filled with ones, and $\odot$ is the element-wise multiplication. The parameter $\lambda$ controls the ratio of combination between the two images and is sampled for each generated sample ($\Tilde{x}$, $\Tilde{y}$) from a beta distribution $\lambda \sim \text{Beta}(\alpha, \alpha)$ defined by the hyper-parameter $\alpha$. In order for M to crop a patch of $x_i$ and replace it with a patch of $x_j$ according to the ratio parameter $\lambda$, a bounding box defined by its center ($C_x$, $C_y$) and its size ($r_w$, $r_h$) is uniformly sampled in the following manner :

\begin{equation}
    \label{eq:bounding_box}
    \begin{split}
        C_w \sim \text{Unif}(0, W),\quad r_w &= W\sqrt{1-\lambda}\\
        C_h \sim \text{Unif}(0, H),\quad r_h &= H\sqrt{1-\lambda} \\
    \end{split}
\end{equation}
Since the area of such a bounding box is $ \frac{r_w r_h}{WH} = 1-\lambda$, filling M with $0$ inside of this box and $1$ elsewhere makes $\lambda$ control the strength of the combination of both the images $(x_i, x_j)$ and labels ($y_i, y_j$).

While the original implementation of CutMix considered mixing samples within the same training mini-batch \cite{yun_cutmix_2019}, we here propose instead to sample $x_i$ from the training mini-batch taken from the incremental dataset $\mathcal{D}_t$ and $x_j$ randomly sampled from the rehearsal memory $\mathcal{M}_t$. We call this variant Rehearsal-CutMix or R-CutMix in short.

$\mathcal{D}_t$ already contains $\mathcal{M}_t$, however, $\mathcal{D}_t$ is biased towards new classes, resulting in mini-batches sampled from it that contain more images from new than past classes. By taking $x_j$ from a second mini-batch sampled solely from $\mathcal{M}_t$, we ensure the generated images are either a new class mixed with a previous class, or two previous classes mixed together which gives three main benefits for our compression step.

On one hand the benefits of the original CutMix augmentation are retained, training on mixed images partly cropped improves the model generalization and robustness \cite{zhang_mixup_2017, devries_improved_2017, yun_cutmix_2019}. On the other hand, specifically in our incremental context, R-CutMix rebalances the training labels and the overall dataset towards past classes which reduces the bias that is learned by the classification layer. Most importantly, it ensures that most training samples contain information about both new and past classes which allows the knowledge distillation loss to transfer better the overall knowledge of the model and especially at the boundaries between each old and new class.

\begin{table*}[h]
    \centering
	\begin{tabular}{l | cccccc | cccc}
		\hline
        \multicolumn{1}{c|}{} & \multicolumn{6}{c|}{CIFAR-100 B0} & \multicolumn{4}{c}{CIFAR-100 B50}\\\cline{2-11} 
		\multicolumn{1}{c|}{} & \multicolumn{2}{c}{5 steps} & \multicolumn{2}{c}{10 steps} & \multicolumn{2}{c|}{20 steps} & \multicolumn{2}{c}{5 steps} & \multicolumn{2}{c}{10 steps} \\ \cline{2-11}
		\multicolumn{1}{c|}{\multirow{-3}{*}{Methods}} & Avg & Last & Avg & Last & Avg & Last & Avg & Last & Avg & Last\\ \hline
		Joint Training & 80.41 & - & 81.49 & - & 81.74 & - & 79.89 & - & 79.91 & -  \\ \hline 
		iCaRL \cite{rebuffi_icarl_2017} & 71.14 & - & 65.27 & 50.74 & 61.20 & 43.75 & 65.06 & - & 58.59 & -  \\
		BiC \cite{wu_large_2019} & 73.10 & - & 68.80 & 53.54 & 66.48 & 47.02 & 66.62 & - & 60.25 & -  \\
		WA \cite{zhao_maintaining_2020} & 72.81 & - & 69.46 & 53.78 & 67.33 & 47.31 & 64.01 & - & 57.86 & -  \\
		DER \cite{yan__2021}& 75.55 & - & 74.64 & 64.35 & 73.98 & \cellcolor{gray!50}62.55 & 72.60 & - & \cellcolor{gray!25}72.45 & - \\
        DER w/o P \cite{yan__2021} & 76.80 & - & 75.36 & 65.22 & 74.09 & \cellcolor{gray!25}62.48 & 73.21 & - & \cellcolor{gray!50}72.81 & - \\
        \textcolor{gray}{DyTox \cite{douillard_dytox_2022}}& \textcolor{gray}{-} & \textcolor{gray}{-} & \textcolor{gray}{73.66} & \textcolor{gray}{60.67} & \textcolor{gray}{72.27} & \textcolor{gray}{56.32} & \textcolor{gray}{-} & \textcolor{gray}{-} & \textcolor{gray}{-} & \textcolor{gray}{-} \\
        \textcolor{gray}{DyTox+ \cite{douillard_dytox_2022}}& \textcolor{gray}{-} & \textcolor{gray}{-} & \textcolor{gray}{75.54} & \textcolor{gray}{62.06} & \textcolor{gray}{75.04} & \textcolor{gray}{60.03} & \textcolor{gray}{-} & \textcolor{gray}{-} & \textcolor{gray}{-} & \textcolor{gray}{-} \\
        FOSTER B4 \cite{wang_foster_2022}& 78.01 & 68.96 & 75.74 & 62.03 & 73.04 & 57.3 & 74.14 & 66.14 & 69.82 & 58.9 \\
        FOSTER \cite{wang_foster_2022}& 77.47 & 68.68 & 75.19 & 62.08 & 72.36 & 57.67 & \cellcolor{gray!25}74.61 & 66.36 & 70.21 & 59.07 \\\hline
        FECIL B4 (ours) & \cellcolor{gray!50}79.16 & \cellcolor{gray!50}71.52 & \cellcolor{gray!50}78.48 & \cellcolor{gray!50}67.31 & \cellcolor{gray!50}76.12 & 61.51 & \cellcolor{gray!50}75.48 & \cellcolor{gray!50}68.23 & 71.18 & \cellcolor{gray!50}61.3 \\
		FECIL (ours) & \cellcolor{gray!25}78.32 & \cellcolor{gray!25}69.35 & \cellcolor{gray!25}77.49 & \cellcolor{gray!25}66.1 & \cellcolor{gray!25}75.0 & 60.41 & 73.99 & \cellcolor{gray!25}66.53 & 70.07 & \cellcolor{gray!25}60.63 \\\hline
	\end{tabular}
	\caption{Results on the CIFAR-100 B0 and B50 benchmarks averaged over three different class orders. The Best and second best methods are displayed respectively with gray and light gray background color. FOSTER was run with the official released implementation, changing only the backbone architecture to a 18-layers ResNet (which increases performance) for fair comparison with other methods. Dytox \cite{douillard_dytox_2022} performances are grayed out because it is the only method that cannot be run with a 18-layer ResNet and instead uses a Vision Transformer.}
	\label{tab:sota_cifar100}
\end{table*}

\section{Experiments}
\label{sec:experiments}
This section provides an extensive evaluation of our approach within three datasets that are widely used in class incremental learning, namely, CIFAR-100, ImageNet-100, and ImageNet-1000. We first present the details of our implementation and the hyper-parameters used for our experiments and then compare the performance of our method against several \sota algorithms (see sections \ref{sec:eval_cifar} and \ref{sec:eval_imagenet}) using popular evaluation protocols and incremental datasets. Finally, in section \ref{sec:ablation}, we conduct ablation studies so as to assess the contribution of each component of our method.

\subsection{Experimental Setup and Implementation Details}
	
\paragraph{Datasets and protocols}
The CIFAR-100 dataset \cite{krizhevsky_learning_2009} is composed of 32x32 pixel color images representing 100 classes with 500 training images and 100 evaluation images for each class.
The ImageNet-1000 dataset \cite{russakovsky_imagenet_2015} on the other hand is a large scale dataset containing 1.2 million training images and 50,000 validation images representing 1000 different categories.
Lastly, the ImageNet-100 dataset consists of a subset of the large scale ImageNet-1000 dataset containing only 100 randomly sampled classes.
Following standard practice, we validate the proposed method on CIFAR-100 and ImageNet-100 with two widely used \cite{yan__2021, wang_foster_2022, douillard_dytox_2022} protocols :

\textbf{B0} (base 0) : In this protocol, all the classes of the dataset are separated equally in 5, 10 or 20 incremental steps and the model uses a rehearsal memory of 2000 exemplars.

\textbf{B50} (base 50) : The model is first trained on 50 initial classes, the remaining 50 classes are then separated equally in 5 or 10 incremental steps and the model is trained with a memory of 20 samples per class in each incremental step.

We compare the top-1 accuracy obtained after the last incremental step as well as the average incremental accuracy as defined in \cite{rebuffi_icarl_2017} for all methods trained with these protocols.

For ImageNet-1000, we validate our approach with the protocol known as \textbf{ImageNet-1000 B0} \cite{rebuffi_icarl_2017, yan__2021, wang_foster_2022} that trains the model on all 1000 classes in 10 incremental steps of 100 classes using a memory size of 20000 samples. Moreover, following \cite{wang_foster_2022}, we report the top-1 and top-5 last step and average incremental accuracies on this protocol.

\begin{figure*}[!t]
    \centering
    \includegraphics[width=\textwidth]{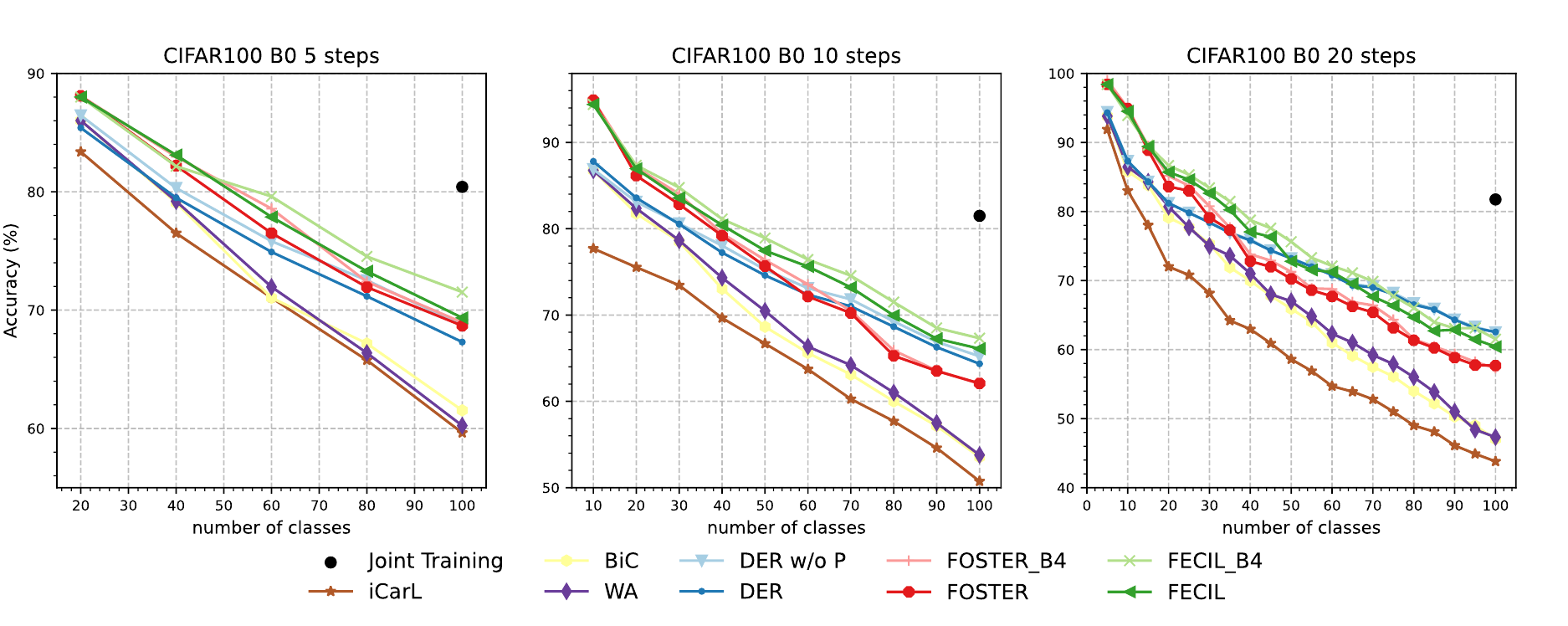}
    \caption{Performance evolution on CIFAR-100. The top-1 accuracy (\%) is reported after each incremental step. Left is evaluated with 5 steps, middle with 10 steps, and right with 20 steps.}
    \label{fig:curves_cifar100}
\end{figure*}

\paragraph{Implementation details}
Our method is implemented in PyTorch \cite{paszke_automatic_2017} with the framework PyCIL \cite{zhou_pycil_2022}. Following~\cite{yan__2021, douillard_dytox_2022} we chose the 18-layers ResNet~\cite{he_deep_2015} backbone architecture to evaluate our model on all the considered datasets.

We note that some previous works \cite{rebuffi_icarl_2017, wu_large_2019, zhao_maintaining_2020, wang_foster_2022} instead used a modified 32-layers ResNet \cite{rebuffi_icarl_2017} for the CIFAR-100 dataset which has been shown in \cite{yan__2021} to underestimate their performance because it cannot achieve competitive results on CIFAR100 compared with the standard 18-layers ResNet \cite{yan__2021}. Authors from \cite{yan__2021}, re-implemented or used the officially released implementations when possible for all the baseline methods \cite{rebuffi_icarl_2017, wu_large_2019, zhao_maintaining_2020} to obtain their performances with a standard 18-layers ResNet architecture. We therefore follow \cite{douillard_dytox_2022} and report these published results in our tables, and use the official FOSTER implementation to compare all methods using a the same 18-layers ResNet model.

Following standard practice, we used the optimizer SGD with a momentum of 0.9, a weight decay of 0.0005 and use a batch size of 128 for CIFAR-100 and 256 for ImageNet. During each training step, our models where trained for 200 epochs, with a learning rate starting at 0.1 and gradually decaying to 0 with a cosine annealing schedule \cite{loshchilov_sgdr_2017}. Following \cite{wang_foster_2022}, the data augmentation applied to training images consists in random cropping, horizontal flip, AutoAugment \cite{cubuk_autoaugment_2019}, and normalization. For the compression phase we add our rehearsal-CutMix augmentation, set the beta distribution parameter $\alpha$ to $0.2$, and set $\tau$ to 2 for the distillation loss. Finally, the exemplars stored in rehearsal memory are selected via the Herding selection strategy \cite{welling_herding_2009} following previous works \cite{rebuffi_icarl_2017}.

\subsection{Evaluation on CIFAR-100}
\label{sec:eval_cifar}

In order to properly evaluate the effectiveness of our method we compare it to several \sota methods including other dynamic network based approaches such as DER \cite{yan__2021}, FOSTER \cite{wang_foster_2022}, and Dytox \cite{douillard_dytox_2022}, and several methods that do not rely on dynamic networks such as iCarL \cite{rebuffi_icarl_2017}, BiC \cite{wu_large_2019}, and WA \cite{zhao_maintaining_2020}.

DER uses a pruning mechanism to ensure the number of parameters does not grow too much and DER w/o P denotes the performance obtained without the pruning part. Our method and FOSTER, on the other hand, use a compression step instead and we therefore denote by FECIL B4 and FOSTER B4 the performances obtained before the compression step. Finally, the method Dytox is the only method that does not use a 18-layers ResNet model and instead leverage a transformer architecture with a similar number of parameters that allows them to control the expansion of parameters during the incremental training.

Table \ref{tab:sota_cifar100} summarizes the main results obtained on the CIFAR-100 dataset. As can be seen in this table our method surpasses significantly the other methods in most experimental settings. When compared to the other expansion and compression-based method FOSTER, our method generally achieves both a higher average incremental accuracy and accuracy after the last step, which demonstrates the effectiveness of our R-mixup compression step. Particularly under the incremental setting B0 10 steps, our FECIL model surpasses FOSTER by $2.3\%$ points of accuracy in average and up to $4\%$ in the last incremental step, demonstrating that our R-CutMix compression scales better with the number of incremental steps.

For the B0 benchmark, as can be seen in the figure \ref{fig:curves_cifar100} our method consistently reaches a higher accuracy than FOSTER and FOSTER B4 after each incremental steps. In fact, only DER reaches a slightly higher accuracy in the last few incremental steps of the B0 20 steps protocol, which is to be expected because DER does not compress the model after each incremental step and instead considers a growing number of parameters, thus allowing their performance to scale better for very high numbers of incremental steps.

Results obtained on the CIFAR-100 B50 protocol are displayed in the Table \ref{tab:sota_cifar100}, where it can be observed that our method still surpass other methods in most cases.

There are two main differences between the B0 and B50 protocols; on the one hand, since half of the dataset is trained initially, remembering the initial classes is much more important for the B50 setting. On the other hand, the rehearsal memory is limited to 20 exemplars per class instead of the total size being limited to 2000 for the B0 protocol. This low number of samples stored in memory is especially challenging for our method as it directly impacts the variability of images sampled by our R-mixup procedure. For this reason our method does not reach the performance of DER in the B50 10steps setting but still outperforms FOSTER, demonstrating the effectiveness of our compression step even with such a limited amount of samples stored in memory. Moreover, with the B50 5steps protocol, the memory grows faster to 2000 samples which positively impacts the performance of our R-CutMix compression and thus allows our method to surpasses both FOSTER and DER.
\begin{table*}[t]
    \centering
	\begin{tabular}{l | c | cc | cc | c | cc | cc}
		\hline
	    & \multicolumn{5}{c |}{ImageNet-100 10 steps} & \multicolumn{5}{c}{ImageNet-1000 10 steps} \\ \cline{2-11} 
		& & \multicolumn{2}{c}{top-1} & \multicolumn{2}{c |}{top-5} & & \multicolumn{2}{c}{top-1} & \multicolumn{2}{c}{top-5} \\ \cline{3-6} \cline{8-11} 
		\multicolumn{1}{c|}{\multirow{-3}{*}{Methods}} & \multicolumn{1}{c|}{\multirow{-2}{*}{\#Params}} & \multicolumn{1}{c}{Avg} & \multicolumn{1}{c}{Last} & \multicolumn{1}{c}{Avg} & \multicolumn{1}{c |}{Last} & \multicolumn{1}{c|}{\multirow{-2}{*}{\#Params}} & \multicolumn{1}{c}{Avg} & \multicolumn{1}{c}{Last} & \multicolumn{1}{c}{Avg} & \multicolumn{1}{c}{Last} \\ \hline
		Joint Training & 11.22 & - & 81.20 & - & 95.1 & 11.68 & - & - & - & 89.27  \\ \hline 
		iCaRL \cite{rebuffi_icarl_2017} & 11.22 & - & - & 83.6 & 63.8 & 11.68 & 38.4 & 22.7 & 63.7 & 44.0 \\
		BiC \cite{wu_large_2019} & 11.22 & - & - & 90.6 & 84.4 & 11.68 & 62.73 & 50.1 & 83.80 & 72.70 \\
		WA \cite{zhao_maintaining_2020} & 11.22 & - & - & 91.0 & 84.1 & 11.68 & 65.67 & 55.60 & 86.6 & 81.1 \\
		DER w/o P \cite{yan__2021} & 112.27 & 77.18 & 66.70 & 93.23 & 87.52 & 116.89 & 68.84 & 60.16 & 88.17 & \cellcolor{gray!25}82.86 \\ 
        DER \cite{yan__2021} & - & 76.12 & 66.06 & 92.79 & 88.38 & - & 66.73 & 58.62 & 87.08 & 81.89 \\
        \textcolor{gray}{DyTox \cite{douillard_dytox_2022}} & \textcolor{gray}{11.01} & \textcolor{gray}{77.15} & \textcolor{gray}{69.10} & \textcolor{gray}{92.04} & \textcolor{gray}{87.98} & \textcolor{gray}{11.36} & \cellcolor{gray!50}\textcolor{gray}{71.29} & \cellcolor{gray!50}\textcolor{gray}{63.34} & \textcolor{gray}{88.59} & \textcolor{gray}{84.49} \\
        FOSTER B4 \cite{wang_foster_2022} & 22.44 & 76.54 & 67.08 & 93.12 & 88.89 & 23.36 & 68.34 & 58.53 & 89.18 & 81.77 \\
        FOSTER \cite{wang_foster_2022} & 11.22 & 76.22 & 66.70 & 93.08 & 88.56 & 11.68 & 68.29 & 58.16 & 89.22 & 81.49 \\\hline
        FECIL B4 (ours) & 22.44 & \cellcolor{gray!50}78.78 & \cellcolor{gray!50}69.25 & \cellcolor{gray!50}95.34 & \cellcolor{gray!50}91.31 & 23.36 & \cellcolor{gray!25}70.05 & \cellcolor{gray!25}61.23 & \cellcolor{gray!50}91.37 & \cellcolor{gray!50}83.19 \\
		FECIL (ours) & 11.22 & \cellcolor{gray!25}78.07 & \cellcolor{gray!25}68.41 & \cellcolor{gray!25}94.76 & \cellcolor{gray!25}90.28 & 11.68 & 69.11 & 60.4 & \cellcolor{gray!25}90.18 & 81.72 \\ \hline
	\end{tabular}
	\caption{Results on the ImageNet-100 and ImageNet-1000 B0 benchmarks. We report the average and last top-1 and top-5 accuracy obtained and display the Best and second best methods respectively with gray and light gray background color.``\#Params'' corresponds to the number of parameters used by the model at the end of the incremental training (in millions).}
	\label{tab:sota_imagenet1000}
\end{table*}

\subsection{Evaluation on ImageNet}
\label{sec:eval_imagenet}

Table \ref{tab:sota_imagenet1000} summarizes the main results obtained for all approaches on the ImageNet-100 and ImageNet-1000 datasets. Similarly to the results obtained on CIFAR-100 with the B0 protocol, both FECIL B4 and FECIL reach higher average and last step accuracy than FOSTER B4 and FOSTER demonstrating the effectiveness of our R-CutMix compression across several datasets. 

On the ImageNet-100 dataset it can be seen that our method consistently reaches significantly higher performance than most other methods on both datasets. Specifically, FECIL B4 and FECIL outperform FOSTER B4 and FOSTER by $2.24\%$ and $1.85\%$ in terms of average top-1 accuracy and $2.17\%$ and $1.71\%$ in terms of last step top-1 accuracy.

Finally, On the Imagenet-1000 dataset it can be observed that our method performs better only in terms of top-5 accuracy. In fact, Dytox reaches significantly higher top-1 performance, however our method reaches higher average and last step top-1 accuracy than DER w/o P while using approximately 10 times less parameters due to our compression step.

\subsection{Detailed analysis of the method}
\label{sec:ablation}

We further evaluate our method and the contribution of each specific component by conducting a thorough ablation analysis of our method. These ablation experiments were conducted on the 10 steps CIFAR-100 B0 benchmark following standard practice \cite{yan__2021}.

Specifically, our expansion training step is very similar to the one exhaustively studied in DER\cite{yan__2021} apart from the fact that it is followed by a compression step. We therefore focus our analysis on the effectiveness of the different components of our compression step. Specifically, we first remove completely our R-CutMix augmentation and perform a naive distillation-based compression after each incremental step. We then compare the effect of the addition of both the standard Mixup \cite{zhang_mixup_2017} and CutMix \cite{yun_cutmix_2019} augmentations. Finally, we replace these augmentations with the rehearsal-based variant used in our method and explained in detail in section \ref{subsec:compression}.

As can be seen in Table \ref{tab:ablations_fecil}, when the standard mixup augmentation is used, the performance attained by our FECIL model after the last incremental step only goes from $61.59\%$ to $61.81\%$ while going up to $62.98\%$ with the rehearsal variant. This demonstrates the benefits of training with mixed images representing samples partly from old and new classes. Furthermore, while even the regular CutMix augmentation increases performance significantly, the gap between CutMix and R-CutMix is much larger than between Mixup and R-Mixup. In fact, while replacing Mixup with R-Mixup improves accuracy reached by our compressed model by $1.17\%$, replacing CutMix with R-CutMix improves the accuracy of the compressed model by $2.88\%$.

\paragraph{Computational overhead} We further evaluate the overhead induced by our R-CutMix procedure by comparing the average time taken by each ablation to accomplish $1$ epoch of the compression step in the table \ref{tab:ablations_fecil}. These times are normalized so that the fully ablated method that took $6.55$ seconds on our computer corresponds to $1x$ and the increase in percentage is noted for the different ablations. It can be observed that the training time overhead induced by the regular mixup and CutMix methods is negligible, however, it is not the case for the rehearsal variants. This increase comes from the fact that for every mini-batch sampled from the dataset a second one is sampled from the rehearsal memory. These two batches, however, are mixed and merged back into a singular mini-batch before going into the model. This prevents any impact on the time needed for the forward and backward pass and therefore explains why the training time overhead remains limited.
\begin{table*}[t]
\centering
\begin{tabular}{c|ccc|cc|cc|c}
    & \multirow{2}{*}{\makebox[1.5em]{\rotatebox{60}{\footnotesize Mixup}}} & \multirow{2}{*}{\makebox[1.5em]{\rotatebox{60}{\footnotesize CutMix}}} & \multirow{2}{*}{\makebox[1.5em]{\rotatebox{60}{\footnotesize Rehearsal}}} & \multicolumn{2}{c|}{FECIL B4} & \multicolumn{2}{c|}{FECIL}& Avg\\
    & & & & Avg & Last & Avg & Last & Time/epoch\\ \hline
    \multirow{5}{*}{\rotatebox{90}{\textbf{Ablations}}} & & & & 76.12 & 63.55 & 75.07 & 61.59 & 1x\\
    & \ding{51} & & & 76.89 & 64.53 & 75.01 & 61.81 & 1.01x \\
    & & \ding{51} & & 77.83 & 66.04 & 76.19 & 63.22 & 1.07x \\
    & \ding{51} & & \ding{51}& 77.48 & 64.47 & 76.35 & 62.98 & 1.22x\\
    & & \ding{51} & \ding{51}& 78.49 & 67.31 & 77.49 & 66.1 & 1.24x\\
\end{tabular}
\caption[Ablation study of our FECIL algorithm.]{Ablation of different key components of our method. The average and last step accuracy are reported for each ablation. We also normalize and report the average time taken by one epoch during the compression step. All experiments were done on the CIFAR-100 B0 10 steps benchmark.}
\label{tab:ablations_fecil}
\end{table*}
\begin{figure}[h]
    \centering
    \includegraphics[width=.5\textwidth]{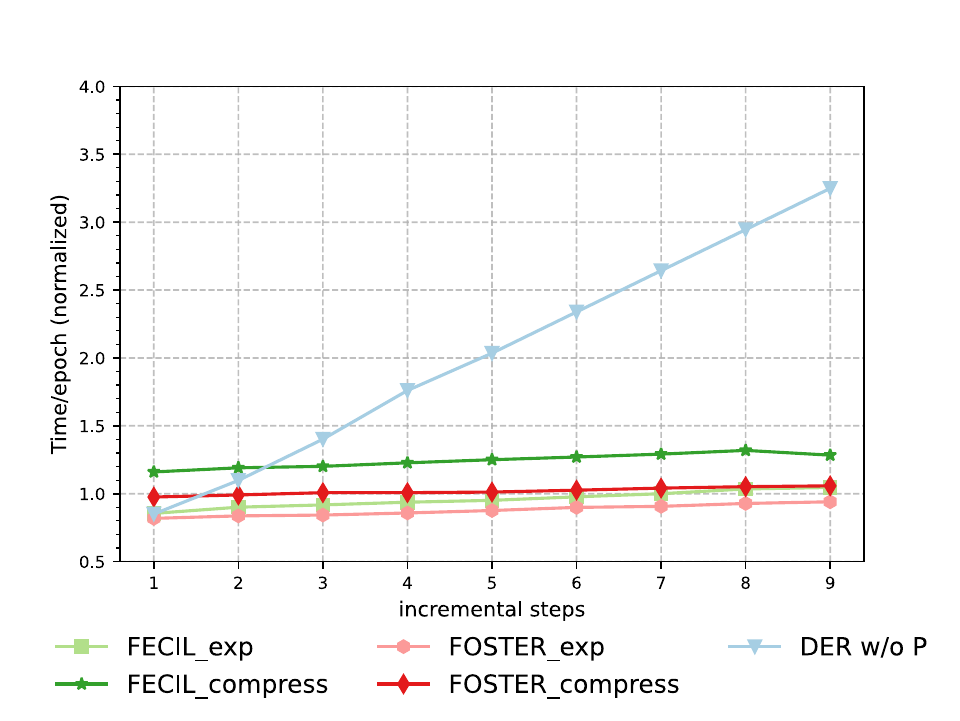}
    \caption{Evolution of the time necessary for an epoch during 10 incremental steps on the CIFAR-100 dataset. FOSTER\_exp and FECIL\_exp represent the time per epoch during the expansion phase while FOSTER\_compress and FECIL\_compress illustrate the time per epoch of the compression step.}
    \label{fig:time_perfs}
\end{figure}

Finally, we compare the training time overhead of our method against FOSTER and DER w/o pruning with the CIFAR-100 B0 10 step protocol in figure \ref{fig:time_perfs}. The times reported are kept normalized in a similar manner than in table \ref{tab:ablations_fecil} for better readability. As can be observed, since DER w/o pruning does not use a compression step, the time necessary for an epoch drastically increases over the incremental steps. Conversely, our approach introduces a minor overhead in contrast to FOSTER, but one that remains constant over time. This ensures that the epoch duration of FECIL remains stable throughout incremental training, thus preserving scalability even with a high number of incremental steps.

\section{Conclusion}
\label{sec:conclusion}
In this work, we introduce FECIL, a novel two-stage training procedure for class incremental learning. Our method first proceeds by expanding the features of the model in order to accommodate new classes without forgetting past ones; It then compresses it back to its original size in order to keep the model size fixed over the course of the entire incremental training process. Specifically, we introduce a Rehearsal-CutMix data augmentation that mixes training images of new classes with images from the rehearsal memory so as to greatly improve the distillation of past classes' information during the compression step. Extensive experiments were performed on three major incremental datasets and a variety of evaluation protocols where our method consistently outperformed other \sota methods. 

While the experiments demonstrated the effectiveness of our compression step, its performance remains bounded by the accuracy reached in the expansion step. As this expansion step is done on an imbalanced dataset, we believe there is still room for further improvements. For example, studying the addition of other Mixup or CutMix-based data augmentations specifically designed to reduce the bias learned in this expansion step could be a promising research direction.

\bibliographystyle{elsarticle-num-names} 
\bibliography{biblio.bib}

\end{document}